\documentclass[conference]{IEEEtran}
\IEEEoverridecommandlockouts

\usepackage[sorting=none, backend=bibtex, giveninits=true]{biblatex}
\usepackage{amsmath,amssymb,amsfonts}
\usepackage{algorithmic}
\usepackage{graphicx}
\usepackage{textcomp}
\usepackage{xspace}

\usepackage{subcaption}
\usepackage{bm}
\usepackage{tabularx}
\usepackage{multirow}
\usepackage{threeparttable}
\usepackage{url}
\usepackage{pgfplots}
\usepackage{placeins}
\graphicspath{{./figures/}}
\usepackage[hidelinks]{hyperref}

\newcommand{\etal}{\xspace{}et al.\xspace}

\newcommand{\reffig}[1]{Fig.~\ref{#1}}
\newcommand{\reftab}[1]{Table~\ref{#1}}
\newcommand{\refsec}[1]{Section~\ref{#1}}

\bibliography{root}

\usepackage{eso-pic}

\AtBeginDocument{\AddToShipoutPictureFG*{\AtTextUpperLeft{\put(0,\LenToUnit{10pt}){\parbox{\textwidth}{\centering\bfseries International Conference on Robotic Computing (IRC), Taichung, Taiwan, 2021
}}}}}

\begin{document}

\title{Flexible-Joint Manipulator Trajectory Tracking with Learned Two-Stage Model employing One-Step Future Prediction\\
\thanks{This work was funded by grant BE 2556/16-2 (Research Unit FOR 2535 Anticipating Human Behavior) of the German Research Foundation (DFG).}
}

\author{
	\IEEEauthorblockN{Dmytro Pavlichenko}
	\IEEEauthorblockA{
		Autonomous Intelligent Systems\\
		University of Bonn, Germany\\
		Email: {\tt pavlichenko@ais.uni-bonn.de}
	}
\and
	\IEEEauthorblockN{Sven Behnke}
	\IEEEauthorblockA{
		Autonomous Intelligent Systems\\
		University of Bonn, Germany\\
		Email: {\tt behnke@cs.uni-bonn.de}
	}
}

\maketitle

\begin{abstract}
Flexible-joint manipulators are frequently used for increased safety during human-robot collaboration and shared workspace tasks. However, joint flexibility significantly reduces the accuracy of motion, especially at high velocities and with inexpensive actuators. In this paper, we present a learning-based approach to identify the unknown dynamics of a flexible-joint manipulator and improve the trajectory tracking at high velocities. We propose a two-stage model which is composed of a one-step forward dynamics future predictor and an inverse dynamics estimator. The second part is based on linear time-invariant dynamical operators to approximate the feed-forward joint position and velocity commands. We train the model end-to-end on real-world data and evaluate it on the Baxter robot. Our experiments indicate that augmenting the input with one-step future state prediction improves the performance, compared to the same model without prediction. We compare joint position, joint velocity and end-effector position tracking accuracy against the classical baseline controller and several simpler models.
\end{abstract}

\section{Introduction}
\label{sec:Introduction}

Robot manipulators have been used for decades, and trajectory tracking control has been extensively researched to achieve fast and accurate motion. In the case of classical industrial manipulators, approaches like iterative learning control (ILC)~\cite{arimoto_1990} can efficiently achieve this goal. ILC assumes that the exact same trajectories are repeated in a well-structured environment. In recent years, the role of robotic manipulators is extending beyond such scenarios: direct human-robot collaboration in shared workspaces induces significantly increased safety requirements. Frequently, their satisfaction starts at the hardware level by the use of compliant series-elastic actuators. However, flexible manipulators produce less accurate motions and the complex underlying dynamics models are often unknown. The simplest way to address this issue is operation at low velocities. Unfortunately, this limits the efficiency of the system. The objective of this paper is to achieve accurate trajectory execution at high velocities for inexpensive flexible-joint manipulators.

\begin{figure}[t]
	\centering
	\includegraphics[width=8.cm]{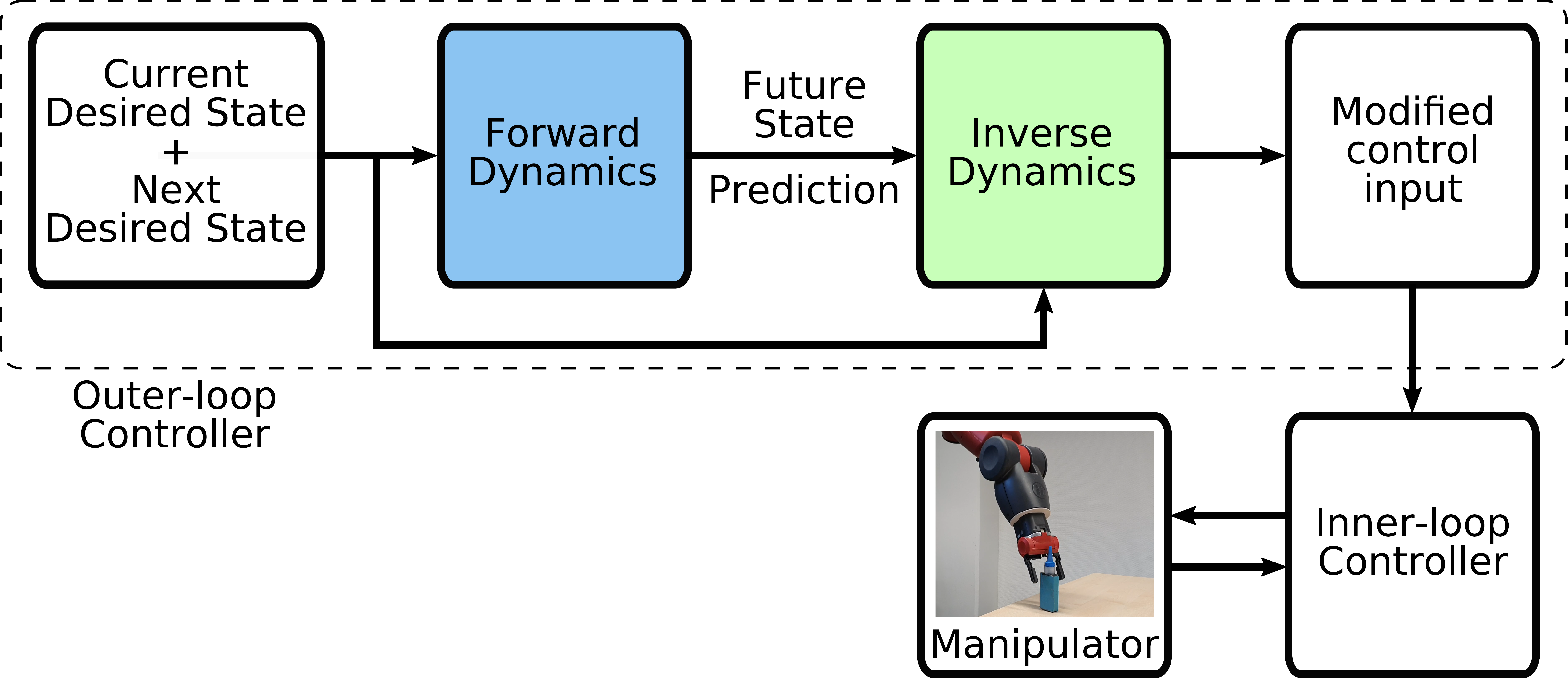}
	\caption{Two-stage model, employing input augmentation with one-step future prediction for inverse dynamics approximation.}
	\label{fig:teaser}
\end{figure}

Neural networks (NNs) are known for their ability to generalize and model complex non-linear relations. We present a methodology for neural-learned feed-forward outer-loop control based on linear time-invariant (LTI) dynamical operators. In particular, as a part of our model we use the novel dynoNet~\cite{piga_2020}, which resembles the features of RNN~\cite{greff_2017} and 1-D convolution~\cite{wang_2017}. The LTI layers are specifically designed for sequence modeling and system identification, successfully approximating complex non-linear causal dynamics, while being differentiable and suitable for backpropagation. Thus, we utilize them for approximating the inverse dynamics of flexible-joint manipulators. We propose a two-stage model (\reffig{fig:teaser}). The motivation for such model architecture is to hardwire the \emph{"Infer what will happen in the future, then think what would be the best action now to prevent the foreseen inaccuracies"} structure within the network. We elaborate that such an architecture moves away from the pure reactive policy towards the more intelligent planning-ahead behavior. This can be viewed as a simplistic model-predictive framework. The first part approximates the forward dynamics of the manipulator. The output of this model is used to augment the input to the LTI-based model, which approximates the inverse dynamics, producing the feed-forward joint position and velocity commands, which are then being fed to the inner-loop feedback controller of the robot manipulator. Explicitly utilizing both forward and inverse dynamics helps to maximize the extraction of information from the scarce real-world data. The models are trained end-to-end on the real-robot data in a supervised manner using plain backpropagation.

We represent each point of the manipulator trajectory as a tuple $\langle \theta, \dot{\theta}, \ddot{\theta} \rangle$, where $\theta$ is a joint positions vector. Explicit inclusion of velocity and acceleration provides information about the dynamics to the NN directly, as opposed to forcing the NN to infer it from the series of joint positions. This state formulation should allow learning the dynamics of the manipulator from a smaller dataset. It is especially important in case of flexible-joint manipulators, which often do not have an accurate dynamics model which could be used to pre-train the model in simulation. Thus, limited real-world data should be used for training. We evaluate the performance of our method on the real Baxter robot against the baseline controller, a multi-layer fully-connected NN, RNN, and plain dynoNet without future prediction step. The training is done on a small real-world dataset. Our approach significantly improves trajectory tracking accuracy, compared to the baseline controller, and outperforms other models. The method allows executing fast trajectories with higher accuracy.

The main contributions of this work are:
\begin{itemize}
	\item Two-stage model architecture with one-step future prediction for feed-forward trajectory tracking, trained end-to-end with backpropagation,
	\item investigation of effectiveness of LTI-based models for robotic manipulator inverse dynamics approximation.
\end{itemize}
\section{Related Work}
\label{sec:Related_Work}

Trajectory tracking has been thoroughly studied for decades. One of the classical methods is iterative learning control~\cite{arimoto_1990} (ILC): it iteratively updates the control input with tracking error and after a small number of iterations is able to perfectly track a desired trajectory. ILCs main limitation is non-transferability: the optimization has to be performed from scratch for each new trajectory. Differential dynamic programming~\cite{mayne1966ASG} (DDP) is another approach, which utilizes a linear quadratic regulator (LQR) in order to iteratively update the control inputs. This approach requires much computational power and is impractical to apply online.

In recent years, trajectory tracking was often addressed by means of NNs~\cite{jiang_2017}~\cite{jin_2018}. Such increased attention is explained by their comprehensive ability to generalize and model complex nonlinear dynamics. Radial basis function (RBF) NNs are a popular model choice~\cite{yang_2016}~\cite{chen_2016}~\cite{han_2016}. Xia \etal~\cite{xia_2014} used RBF-NN to mitigate the effects of friction in swing-up control of a two-joint manipulator. However, such models contain a large number of parameters, and it is a challenging task to tune the hyper-parameters, such as number of Gaussian kernels, their centers and shapes. A wavelet fuzzy neural network is proposed for predictive control by Lu~\cite{lu_2011}. The model is based on a set of fuzzy rules, where each rule is linked to the wavelet function from the consequent rules. The model is trained with backpropagation. The drawback of this architecture is its complexity and computational expensiveness. Gaussian process regression (GPR) is used by Rueckert \etal~\cite{rueckert_2017} to perform kinematic control of a surgical cable-driven manipulator. Saveriano \etal~\cite{saveriano_2017} modeled the residual dynamics using GP-based model together with reinforcement learning. The main drawback of GP-based models, in comparison to neural networks, is that they grow together with the data, thus requiring a lot of resources for evaluation and execution. Mahler \etal~\cite{mahler_2014} demonstrated that LSTM networks can outperform the GP-based method for modeling inverse dynamics. While many works focus on the model architecture, Morse \etal~\cite{morse2020} apply meta-learning to obtain state-dependent loss functions, which demonstrates another viable approach.

\begin{figure*}[ht!]
	\centering
	\includegraphics[width=0.95\linewidth]{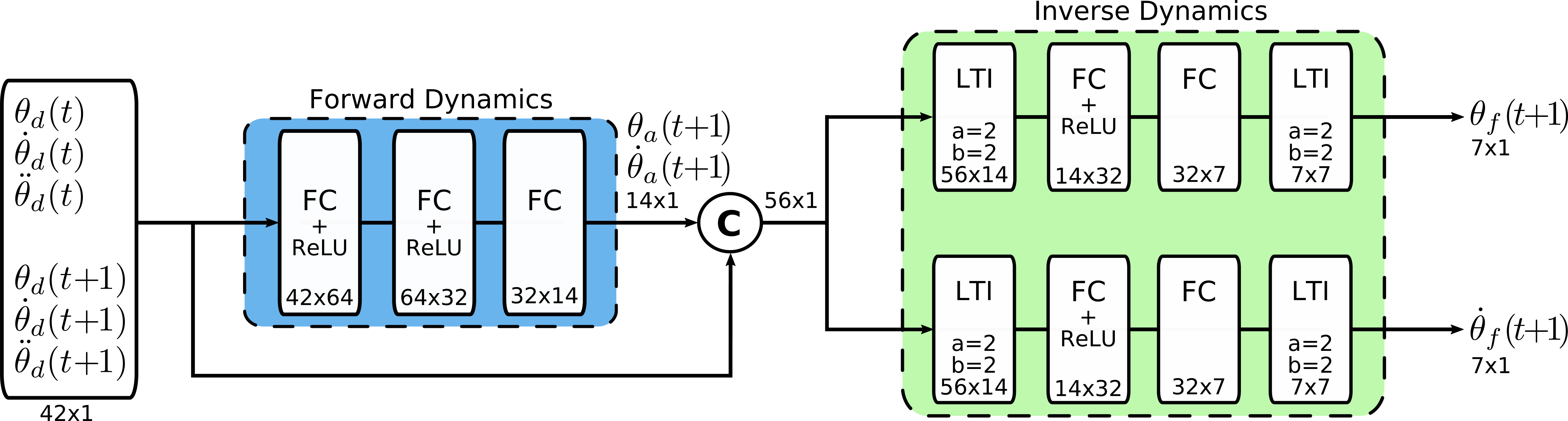}
	\caption{Proposed model. Input is the current desired state of the manipulator (joint position, velocity, acceleration) plus the next desired state. Blue: Forward Inference Network (FIN), which approximates the forward dynamics. Green: dynoNet Wiener-Hammerstein model (DWH), which takes the same input, augmented with the future state estimation from FIN. It outputs the feed-forward joint position and velocity commands. (c): Concatenation. FC: fully-connected layer. LTI: linear time-invariant dynamical operator-based block.}
	\label{fig:model_diagram}
\end{figure*}

Much research is built around ILC~\cite{patan_2017}~\cite{zuo_2010}. For instance, in work by Schwarz \etal~\cite{schwarz_2013} the compliant position control is achieved through learning the parameters of DC motors and friction models with a help of ILC. The method was tested in real world with humanoid robot gait. This approach avoids performing a separate run for each parameter, but instead identifies all parameters at once. A combination of $\mathcal{L}_1$ adaptive control and ILC is used for transfer learning between systems with different dynamics by Pereida \etal~\cite{pereida_2018}. An extended $\mathcal{L}_1$ controller runs in the inner closed-loop control level and achieves robust and repeatable behavior. ILC is used as an outer-loop control, where the transfer of learned experience is realized. The system is tested on two quadrotors with different dynamics and outperforms systems composed of ILC and proportional-derivative (PD) or proportional-integral-derivative (PID) controllers. Another use of ILC involves production of the ground truth input trajectories to train NNs which approximate the inverse dynamics of the system by Chen \etal~\cite{chen2019industrial}. The authors apply this approach to an industrial manipulator, training the model in simulation. The model is then applied to the real robot with transfer learning. A separate NN is trained for each joint using backpropagation. The approach demonstrates a significant improvement of the trajectory tracking accuracy both in simulation and in the real world. However, this method cannot be applied to systems which do not have good dynamics approximation for simulation, which is frequently the case for compliant systems. In addition, the assumption of decoupled joints does not hold when dealing with flexible joints.

Chen \etal~\cite{chen2019neurallearning} present two approaches for performing feed-forward trajectory tracking with series-elastic actuators. The first approach uses RNN which approximates forward dynamics in combination with ILC, which then utilizes this model to find an optimal command sequence. The main drawback of this approach is a significant runtime required for ILC to converge, which makes this method impractical to be applied online. The second approach utilizes bi-directional RNN (BRNN)~\cite{schuster_1997} to approximate the inverse dynamics directly, such as in~\cite{talebi_1998}~\cite{li_2017}. This allows to directly obtain required control inputs in a short time. Both approaches are trained with backpropagation on three hours of sinusoidal and random trajectories recorded on the real Baxter robot. Resulting models improved the trajectory tracking over the baseline PD controller. In contrast, we propose a model which utilizes causal LTI dynamical operators. In addition, we also encapsulate a pre-trained model for forward dynamics approximation to augment the input to the inverse dynamics model. This two-stage approach, inspired by Zeng \etal~\cite{zeng_2020}, allows the model to take into account the future prediction in order to estimate the feed-forward command which mitigates future inaccuracies. Combined with more abstract input representation (two timesteps of joint positions, velocities and accelerations instead of multiple timesteps of joint positions), the proposed model is able to learn and generalize from a smaller amount of real-world data. In addition, our model produces not only feed-forward joint positions commands, but velocities as well, which allows to further improve the trajectory tracking accuracy. Finally, we train the model on a smaller dataset of functional trajectories, emulating learning from a regular operation, as opposed to learning from trajectories of specific artificial shapes, such as sine waves.
\section{Method}
\label{sec:Method}

Given a desired trajectory $\theta_d$ which consists of $T$ equally spaced in time points $\theta_d(t) \in \mathbb{R}^N, t \in [1 ... T]$ in a joint space with $N$ joints, our objective is to produce feed-forward commands $\theta_f$ which would lead the manipulator to follow $\theta_d$. In this section, we present the details on the two-stage model for the manipulator inverse dynamics approximation.

\subsection{Data Collection}
In order to learn the inverse dynamics approximation, we generate 45 minutes of \emph{functional} trajectories $\theta_d$ for the left arm of the Baxter robot. Random sinusoidal trajectories are often used to form the base of the training set~\cite{chen2019industrial}~\cite{chen2019neurallearning}. However, recording of such dataset takes the valuable robot hours away from the user. Thus, we wanted to emulate learning from the data which was collected when performing actual tasks. That is why we refer to such trajectories as \emph{functional}. This is advantageous, because data collection can take place transparently while the robot is performing useful tasks. Subsequently, the learned model extends the range of executable tasks and reduces execution time, improving the overall capabilities of the flexible-joint manipulator.

The training set consists of pick-and-place trajectories with equal portions of them executed with 0.6, 0.8 and 1.0\,rad/s maximum velocities. We did our best to cover the major part of the workspace in front and to the side of the robot. When executing a trajectory $\theta_d$, the actual response of the robot $\theta_a$ and $\dot{\theta}_a$ is recorded. Since Baxter has no way to measure the actual acceleration $\ddot{\theta}_a$, we approximate it by a cubic spline interpolation of the $\dot{\theta}_a$. Approximately 60\% of the trajectories contain 1-2 additional waypoints between the start and the goal. This increases the variety of the movements and simulates maneuvers such as avoiding an obstacle.

\subsection{Two-stage Model}
We propose a two-stage model which consists of two parts (\reffig{fig:model_diagram}). The first part is the Forward-Inference Network (FIN): a fully connected multi-layer NN which approximates manipulator forward dynamics. The second part is based on LTI dynamical operators, as implemented in the novel dynoNet\footnote{\url{https://github.com/forgi86/dynonet}}. LTI operators were shown to be efficient~\cite{piga_2020} when learning the complex causal non-linear dynamics, while being suitable for an end-to-end backpropagation. These properties are advantageous for learning the dynamics of the flexible-joint manipulator. That is why we chose LTI-based blocks to be the core of our model. The model resembles a MIMO Wiener-Hammerstein structure, according to the block-oriented modeling framework~\cite{giri_2010}. Thus, it is referred to as dynoNet Wiener-Hammerstein model (DWH). DWH uses the original input augmented with the FIN prediction to approximate the inverse dynamics. We refer to the whole model as FIN-DWH. The two-stage architecture with one step future prediction aims to push the model from purely reactive policy behavior towards more intelligent planning ahead, which would allow achieving a higher accuracy of the trajectory tracking.

The Baxter arm has $N=7$ joints, thus $\theta(t) \in \mathbb{R}^7$. Since the joints of Baxter are coupled~\cite{chen2019neurallearning}, it is not feasible to train a separate model for each joint. Instead, we approximate the underlying dynamics by considering all joints simultaneously. So, the input to the FIN-DWH model is a 42-element vector:
\begin{equation}
[\theta_d(t), \dot{\theta}_d(t), \ddot{\theta}_d(t), \theta_d(t+1), \dot{\theta}_d(t+1), \ddot{\theta}_d(t+1)],
\end{equation}
where $\theta_d(t)$ is the current desired state of the manipulator and $\theta_d(t+1)$ is the next desired state. The output of the network is then a 14-element vector $[\theta_f(t+1), \dot{\theta}_f(t+1)]$ where $\theta_f(t+1)$ is the feed-forward command which should lead the manipulator to the state $\theta_d(t+1)$. The proposed method is open-loop and does not include live feedback from the robot. There is an assumption that the consequent execution of the corrective feed-forward commands should result in the manipulator following the desired trajectory perfectly. This allows to train offline with the collected data as it is, without the need to introduce an additional dynamics model to produce the ground-truth control inputs, as described in the next subsection. We analyze the practical shortcomings of the aforementioned assumption by performing the experiments with a previously unseen payload. Since the complete feed-forward command sequence from our model is available before it is executed, we apply a zero-phase Savitzky-Golay filter to each individual joint position and joint velocity trajectory with window 21 and polynomial order 2 to alleviate any potential non-smooth fragments in the control signal. 

The best-performing Baxter baseline controller is "Inverse Dynamics Feed Forward
Position control"\footnote{\url{https://sdk.rethinkrobotics.com/wiki/Joint_Trajectory_Action_Server}}. This controller calculates the necessary torque from the supplied positions, velocities and accelerations using the internal dynamics model. Thus, we train our model to produce velocities $\dot{\theta}_f(t+1)$ as well. We do not train the model to output joint accelerations $\ddot{\theta}_f(t+1)$ because they could not be measured by the robot hardware and training with approximated spline accelerations as a target would most likely lead to an inferior performance. To give an intuition about how the position baseline controller and the inverse dynamics feed-forward position baseline controller compare, the average cumulative joint position tracking error per point is $0.157\pm0.082$\,rad in the first case against $0.069\pm0.032$\,rad ($\mu \pm \sigma$) in the second case. The latter is more than two times accurate. Thus, we train the model to provide feed-forward velocity input and compare against this more accurate baseline. Note, that it is straightforward to use only the position or velocity vector from the output in case when position or velocity control is used. The existing error in trajectory tracking accuracy shows that the internal dynamics model does not represent the complex coupled-joints real-robot dynamics accurately enough. The proposed data-driven method does not replace the classical baseline controller, but forms an outer-loop, complementing the existing dynamics model and learning to compensate for the observed inaccuracies.

By explicitly including velocity and acceleration into the input, we provide enriched information about the manipulator state without forcing the network to infer derivatives from the time series of joint positions. In addition, since a tuple of $\langle \theta, \dot{\theta}, \ddot{\theta} \rangle$ contains certain information about the dynamic state of the manipulator, we can significantly reduce the number of time steps needed as an input. In this work, we use only two time steps, as described above. Moreover, velocity and acceleration represent certain patterns in robot dynamics in a more general way than sequences of joint positions alone, which values depend on the specific manipulator position in the workspace. This should allow the model to learn from fewer data points, which is critical when the learning is performed directly on the real-manipulator data. The trajectories used for training contain points separated by $\Delta t=\frac{1}{F}$\,s where $F=20$\,Hz. In case of any encoder inaccuracies of consistent magnitude, a larger time span between sample points allows decreasing their influence. In addition, larger $\Delta t$ also reduces the influence of latency. The inner-loop feedback controller of the Baxter joints operates at a much higher frequency.

For approximating the forward dynamics of the manipulator, we define the FIN model: a simple NN consisting of three fully connected layers. It takes a 42-element vector as an input and produces a 14-element vector $[\theta_a(t+1), \dot{\theta}_a(t+1)]$, where $\theta_a(t+1)$ is a predicted state of the manipulator after executing the command $\theta_d(t+1)$ as it is. We use the ReLU activation function as non-linearity. FIN model has 5,294 weights in total. The following DWH model approximates the inverse dynamics and consists of two identical independent branches. Each branch takes in the original 42-element input concatenated with the 14-element output of FIN model, resulting in a 56-element input. Each branch then outputs a 7-element vector. One branch produces positions, the other -- velocities. The architecture of a branch is as follows: first the LTI block with $a=2$ and $b=2$, which outputs 14 features (motivated to represent a rough position\,+\,velocity approximation). $a$ and $b$ define the polynomial order for the denominator and nominator of a rational transfer function (see~\cite{piga_2020} for details). The LTI block is followed by two fully connected layers. The result of these layers is then fed to the last LTI block which as well has $a=2$ and $b=2$ and produces the final 7-element vector. The overall architecture of the model is shown in \reffig{fig:model_diagram}. Each branch has 4,043 parameters, which results in total of \mbox{4,043$\times2=$ 8,086} parameters for the DWH model. The LTI layers are parameterized in terms of rational transfer functions, and thus apply infinite impulse response (IIR) filtering to the input. Stacking this hardwired linear structure in multiple layers together with fully connected layers was shown~\cite{piga_2020} to approximate the complex non-linear dynamics. By providing the forward dynamics estimation obtained from the FIN, we allow the model to take into account the predicted future in which we would execute the next command as it is. This can be interpreted as a simplistic version of model-predictive control with only one step of looking ahead. As we show in our evaluation, this additional input improves the performance of the model.

\subsection{Training}
Given the set of desired trajectories $\theta_d$, $\dot{\theta}_d$ and $\ddot{\theta}_d$, as well as the set of the actual robot responses $\theta_a$, $\dot{\theta}_a$ and $\ddot{\theta}_a$, we train the two-stage model in two steps.

First, we train the FIN model. Since we have the desired trajectories and the actual responses, the composition of the training input-output tuples is straightforward. We train the network using stochastic gradient descent (SGD) with a minibatch of 32 data points in a fully supervised manner. We use the Adam optimizer with a learning rate of $10^{-4}$ to minimize the mean square error (MSE) loss and employ L2 regularization.

The training of the full FIN-DWH model is not as straightforward, because the ground-truth commands $\theta_f$ are unknown. It would be possible to employ ILC to obtain them, but since we do not have a good model for simulation, this would have to be done on the real robot, significantly increasing the number of robot-hours needed to produce such a dataset. Instead, we use the same data as for FIN training, and apply the Hindsight Experience Replay (HER) technique~\cite{andrychowicz2018hindsight}. This method can be described in short as pretending that what we achieved was what we actually wanted and is commonly used in reinforcement learning to mitigate the negative effects of sparse delayed rewards. We use $\theta_a$ as the goal trajectory and then $\theta_d$ becomes the ground-truth input. After inverting the training examples, we perform the same training procedure as above: 32 data points per minibatch, Adam optimizer with learning rate of $10^{-4}$ with the MSE loss and L2 regularization. Note, that we keep the weights of the FIN frozen during the whole training of the FIN-DWH model. This ensures that the DWH model is supplied with forward dynamics prediction. Moreover, we have observed that allowing the network to update the FIN weights during the training led to inferior performance. This supports our idea that when training on a limited dataset, careful hard-wiring of dataflow structure is of high importance.

Although we train the model directly with trajectories with different velocity profiles, it is possible to train the model by gradually increasing the difficulty. Such training scenario would fit transparently into the real-world application scenario, when allocating robot hours only for training is infeasible and learning from the real tasks is required. In this setting, the tasks would be first executed with low velocities, to minimize the trajectory tracking inaccuracies and allow to successfully complete the tasks. Then, as the model improves the tracking accuracy, the velocity would be gradually increased.
\section{Evaluation and Experiments}
\label{sec:Evaluation}

To evaluate our approach, we conduct experiments on the real Baxter robot. We compare the performance to the baseline Baxter controller (Inverse Dynamics Feed Forward
Position control). In addition, we also compare our method against the three other models. The first model consists of three fully connected layers: ${42 \times 64 \rightarrow 64 \times 32 \rightarrow 32 \times 14}$. It has 5,294 parameters and is further referred to as FC. Note, that we also experimented with a larger FC model with four layers and 16,302 parameters, however, it did not demonstrate a superior performance. The second model is a three-layer RNN, which has 7,772 parameters and is referred to as RNN. It has analogous to FC structure: ${(42 + h) \times 64 \rightarrow 64 \times 32 \rightarrow 32 \times 14}$, with two additional layers to produce a 14-element hidden state $h$: ${(42 + h) \times 32 \rightarrow 32 \times h}$. The third model is dynoNet Wiener-Hammerstein (DWH) with 6,518 parameters. It has the same structure as the network described in \refsec{sec:Method}, except that it does not have a FIN model to estimate the future outcome of feed-forward control. All three models take 42-element vectors as input and produce a 14-element vector of feed-forward joint positions and velocities. All models use ReLU non-linearity and were trained on the same dataset until convergence, minimizing MSE loss with Adam optimizer. We did our best to find the best set of hyper-parameters for each model using grid search. For the FC model they were: learning rate of $2.0 \cdot 10^{-4}$ and minibatch size of 24 data points. For the RNN model they were: learning rate of $1,5 \cdot 10^{-4}$ and minibatch size of 48 data points. For the DWH and FIN-DWH we used the same hyper-parameters (learning rate of $10^{-4}$ and minibatch size of 32 data points), since the core of both models is the same. This allows to better observe how the proposed two-stage architecture influences the performance of the model, compared to the same model without the future forward dynamics prediction step.

\begin{figure}[t]
	\centering
	a)\includegraphics[width=3.9cm]{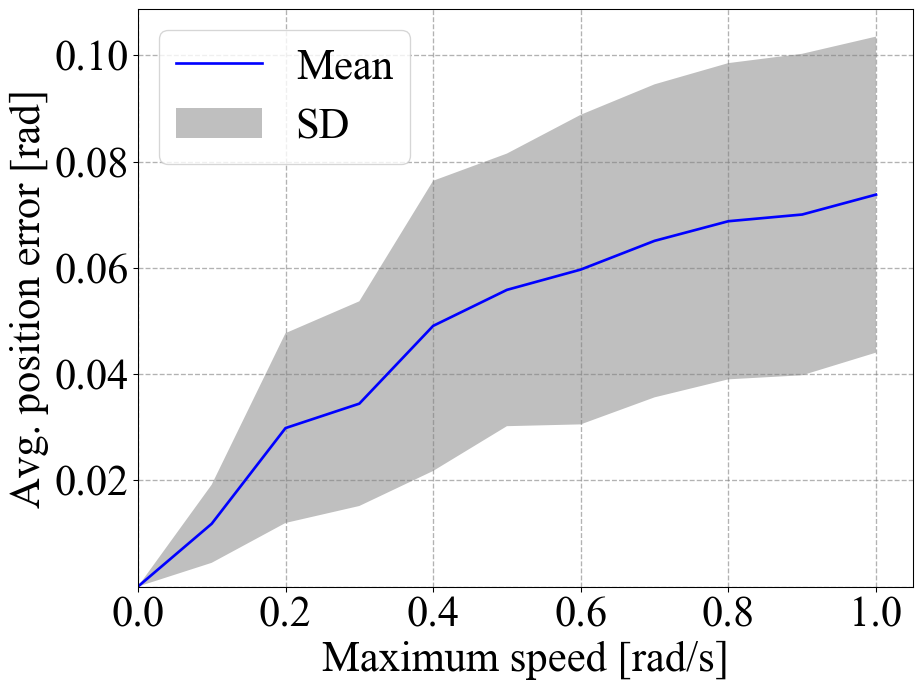}
	b)\includegraphics[width=3.82cm]{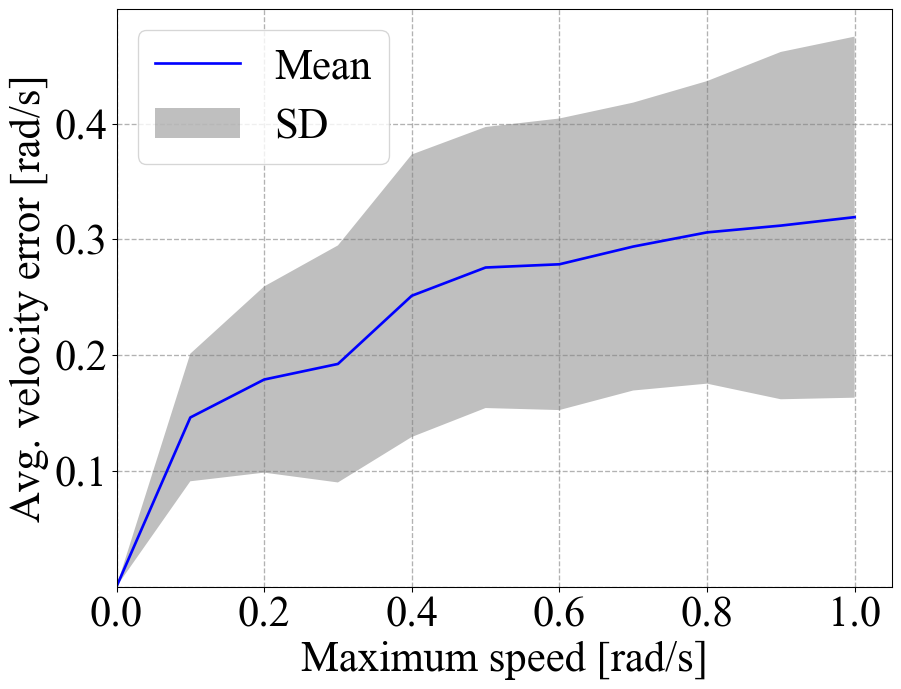}
	\caption{Maximum joint speed in a trajectory vs: a) average position error, b) average velocity error, per trajectory point.}
	\label{fig:error_vs_speed}
\end{figure}

\subsection{Quantitative Evaluation}

To quantitatively evaluate the proposed approach, we generated 100 unseen trajectories for the left arm of Baxter. Although our model can be used for trajectories with arbitrary speed profile, we find it especially interesting to evaluate methods on the fast trajectories because in this case the compliance of the manipulator causes the most inaccuracies due to hard to model inertia effects on flexible joints. In \reffig{fig:error_vs_speed} one can see the plot of maximum allowed speed in a trajectory vs. average cumulative error of joint position and velocity per point in a trajectory executed with a baseline controller. Clearly, the higher the maximum allowed speed is, the less accurate are the movements of the manipulator. The error grows slower with maximum speed increase because not all joints are able to reach this maximum speed during a trajectory. Thus, in our experiments, the maximum joint speed was set to 1.0 rad/s. We measure the average cumulative error per point in a trajectory for the joint position, joint velocity and end-effector position. We also measure the extra time needed to converge to the final point in a trajectory, as well as the runtime of each model. Given an actual trajectory $\theta_a$ and a desired trajectory $\theta_d$ with $T$ points and $N$ joints, we calculate the average cumulative joint position error $e_n$ per point as follows:
\begin{equation}
e_n = \frac{1}{T} \sum_{t=1}^{T} \sum_{n=1}^{N}|\theta_d(t, n) - \theta_a(t, n) |.
\end{equation}
The procedure is analogous for the average joint velocity error and average end-effector position error.

\begin{figure}[t]
	\centering
	\includegraphics[width=4.3cm]{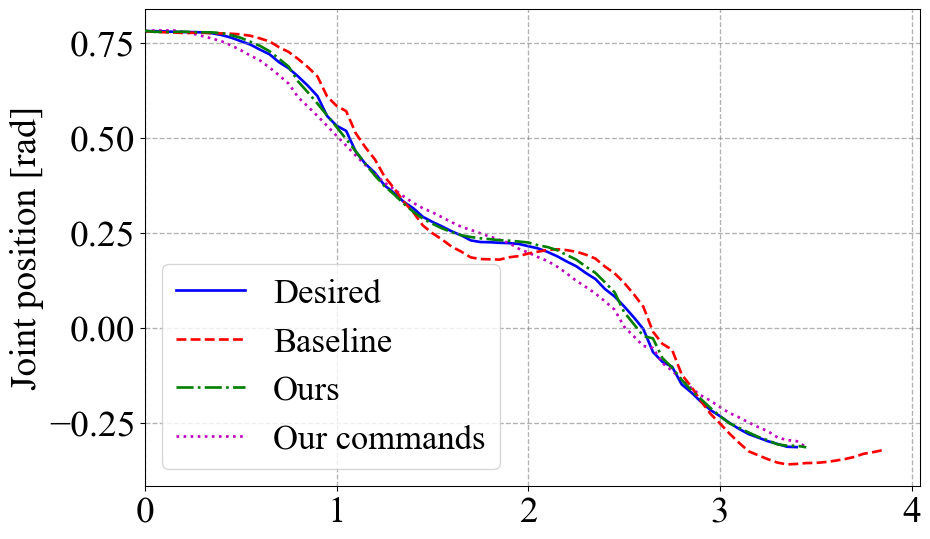}
	\includegraphics[width=4.3cm]{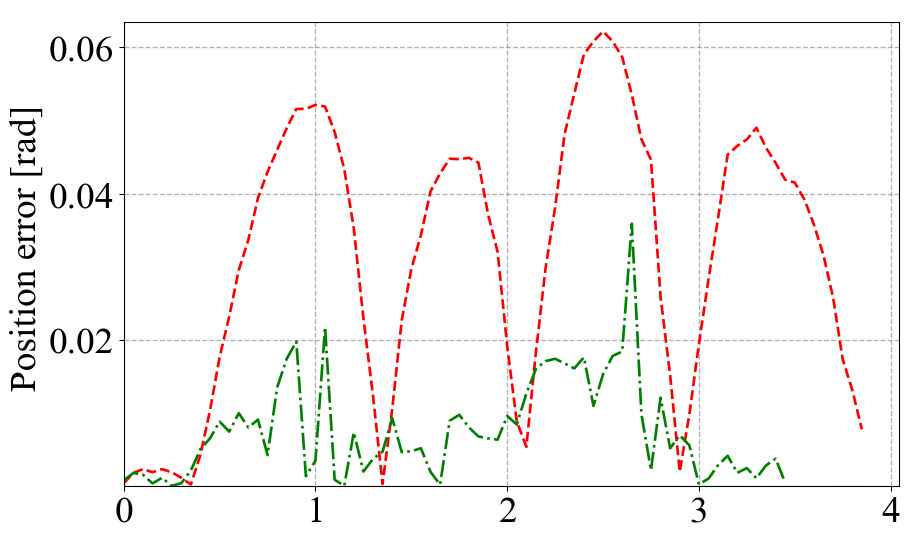}
	\includegraphics[width=4.3cm]{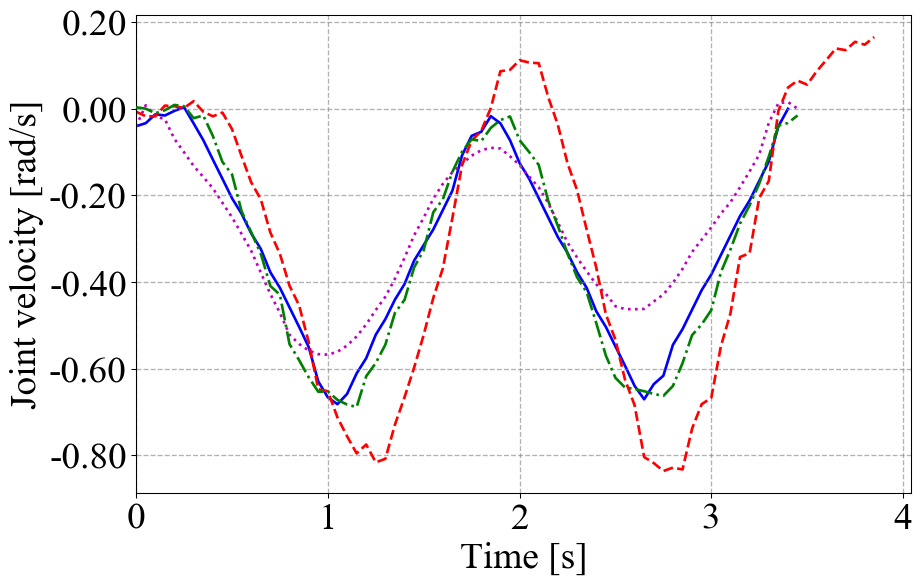} \includegraphics[width=4.3cm]{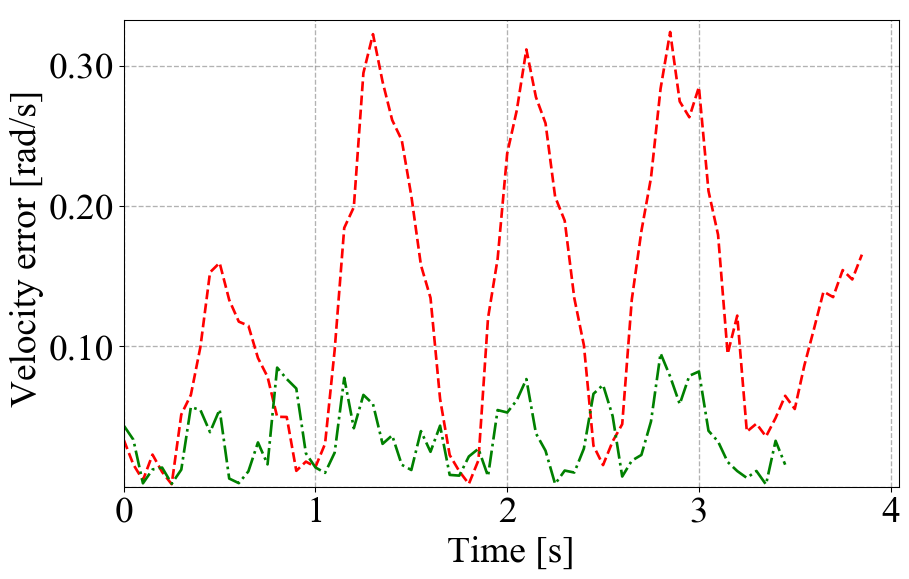}
	\caption{Shoulder yaw trajectory. Top-left: position vs time. Bottom-left: velocity vs time. Top-right: position error vs time. Bottom-right: velocity error vs time. Blue (solid line): desired trajectory. Red (dashed line): execution with baseline controller. Green (dash-dotted line): execution with our method. Magenta (dotted line): feed-forward command of our method. The baseline controller alone leads to several major deviations, while our method allows to compensate for the inertia affecting the elastic joints and follows the trajectory with higher accuracy.}
	\label{fig:joint_pos_vel}
\end{figure}

In \reftab{table:pos_vel_error}, the average cumulative joint position and velocity errors are shown. In all tables we provide 95\% confidence intervals (CI) of the mean. One can see that all the models made an improvement over the sole baseline controller. However, the FC model clearly shows the worst performance. This is the case because raw fully-connected layers do not have an underlying structure to capture the unknown dynamics. RNN and DWH models show quite similar performance, although DWH has slightly better results. Finally, the FIN-DWH improves over the plain DWH, demonstrating that the one-step prediction of the future allows producing a more accurate control input. In \reftab{table:eef_error}, the average end-effector position error per point as well as the extra time to converge to the final point are shown. The extra time is defined as a difference between the actual trajectory execution time and the desired execution time. This difference arises when the state of the manipulator in the final trajectory point is significantly deviated from the desired state and extra time is needed to reach it. A similar tendency on the performance of the models can be observed. On average, the baseline controller has a 3\,cm deviation from the desired path, while our method achieves an improvement of three times, reducing it to 1\,cm on average. The extra time to arrive at the destination point is reduced by significant 92\%, which happens because the motion is much smoother overall, making the immediate accurate arrival at the final point possible. The average runtimes per trajectory (usually consisting of 60-90 time steps) are as follows. FC: 0.031\,s, RNN: 0.062\,s, DWH: 0.067\,s, FIN-DWH: 0.085\,s. All computations were executed on an Intel i7-6700HQ 2.6\,GHz CPU. Further improvements of the runtimes are possible. By listing them here, we give an intuition about the relative computational load of the compared models.

\begin{figure}[t]
	\centering
	\includegraphics[width=7.0cm]{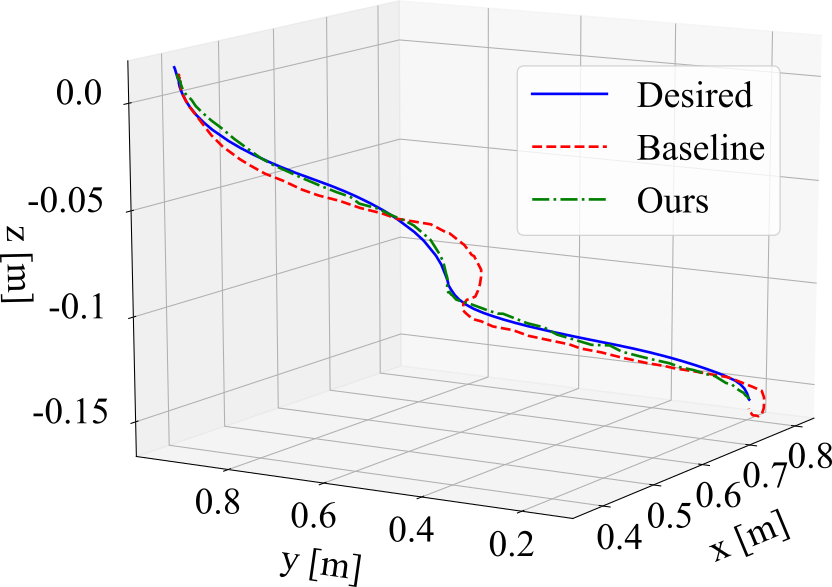}
	
	\vspace{0.5cm}
	
	\includegraphics[width=6.5cm]{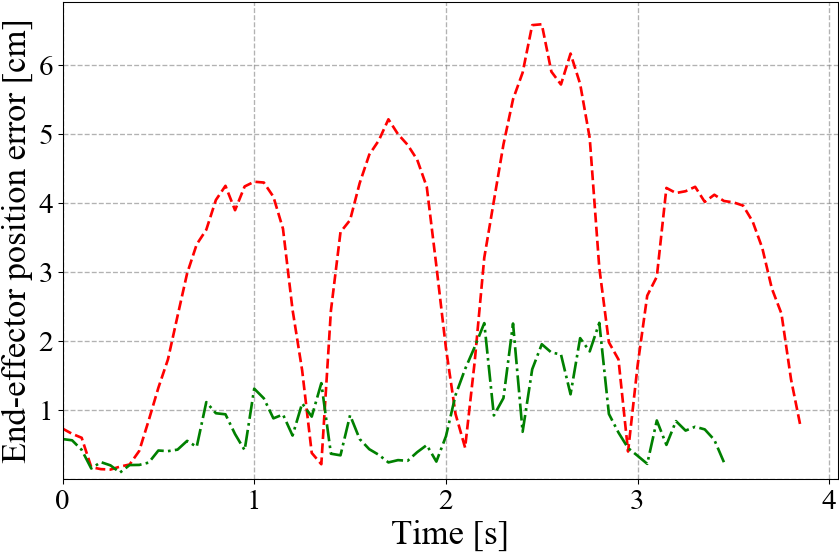}
	\caption{An example trajectory. Top: path of the end-effector in 3D. Bottom: end-effector position error. Blue (solid line): desired trajectory. Red (dashed line): execution with baseline controller. Green (dash-dotted line): execution with our method. Execution with the baseline controller leads to several major deviations, while our approach allows to follow the trajectory with higher accuracy.}
	\label{fig:3d_traj}
\end{figure}

In \reffig{fig:joint_pos_vel}, we show an example test trajectory of the shoulder yaw. One can see that the trajectory achieved solely by the baseline controller deviates from the target much more than the FIN-DWH generated trajectory. There are four big peaks of the position error and three big peaks of the velocity error. It can be noticed that the minima of the position error correspond to the maxima of the velocity error and vice versa. This can be explained by the controller trying to compensate for the position error by increasing velocity to "catch up", however, this consequently leads to the overshoot of the velocity, which is repeated. We attribute this effect to the compliance of the manipulator. It can be seen that our model learned to compensate for this, modifying the joint commands when needed (see the dotted yellow line), achieving a much more precise trajectory tracking of both joint position and velocity. In \reffig{fig:3d_traj}, we show the path and the error of the end-effector position. One can see that with the baseline controller, the end-effector makes several swings, deviating significantly from the desired trajectory. The peaks of the deviations correspond to the peaks of shoulder yaw deviations, because shoulder joints are the ones most affected by inertia, bearing the weight of the whole arm. It is also worth noticing that the baseline trajectory takes around half a second of extra time to reach the final position, while our method requires almost no extra time. Overall, our approach showed better performance than the other three models, and significantly improved over the baseline controller, producing smooth high-velocity trajectories. Achieving higher accuracy over the raw DWH model, the addition of the FIN module was shown to be beneficial.

\begin{table}
	\centering
	\caption{Comparison of average cumulative joint position and velocity error per point.}
	\label{table:pos_vel_error}
	\normalsize
	\begin{tabular}{lcc}
		\hline
		Method &
		\begin{tabular}[c]{@{}c@{}}Joint position\\ error, rad\end{tabular} &
		\begin{tabular}[c]{@{}c@{}}Joint velocity\\ error, rad/s\end{tabular} \\ \hline
		Baseline    & 0.069$\pm$0.0062( --- )  & 0.325$\pm$0.028( --- ) \\
		FC          & 0.047$\pm$0.0061(32\%)   & 0.243$\pm$0.021(25\%) \\
		RNN         & 0.042$\pm$0.0042(39\%)   & 0.227$\pm$0.017(30\%) \\
		DWH         & 0.040$\pm$0.0037(42\%)   & 0.218$\pm$0.016(32\%) \\
		FIN-DWH     & \textbf{0.036$\pm$0.0033(47\%)}   & \textbf{0.213$\pm$0.015(35\%)} \\ \hline
	\end{tabular}
	\begin{tablenotes}
		\item \footnotesize{95\% confidence interval is provided after "$\pm$". Improvement over the baseline is in brackets.}
	\end{tablenotes}
\end{table}

In addition, we execute 20 random trajectories from the previous experiment, while holding a payload of 0.25, 0.5 and 1.1\,kg. Note, that the maximum payload for Baxter is 2.3\,kg. In \reftab{table:payload}, the average cumulative joint position and velocity errors are shown. It is possible to see that while carrying the smallest payload of 0.25\,kg, both the baseline and the proposed FIN-DWH model perform similarly to the run without the payload (\reftab{table:pos_vel_error}). Increasing the payload to 0.5\,kg causes a more noticeable decrease of the trajectory tracking accuracy. Finally, with a significant load of 1.1\,kg, the trajectory tracking accuracy declines substantially, influenced by the inertia forces of larger magnitude. The largest deviations were observed in parts of the trajectories with relatively fast changes in acceleration. The proposed model was trained on payload-free trajectories and does not incorporate live feedback from the robot. Nevertheless, this experiment demonstrates that the learned dynamics model of the manipulator helps to reduce the negative impact of the unaccounted payload and achieve more accurate trajectory execution.

\subsection{Practical Example}

\begin{figure*}[t]
	\centering
	\hspace{1.8ex} \includegraphics[width=3.0cm]{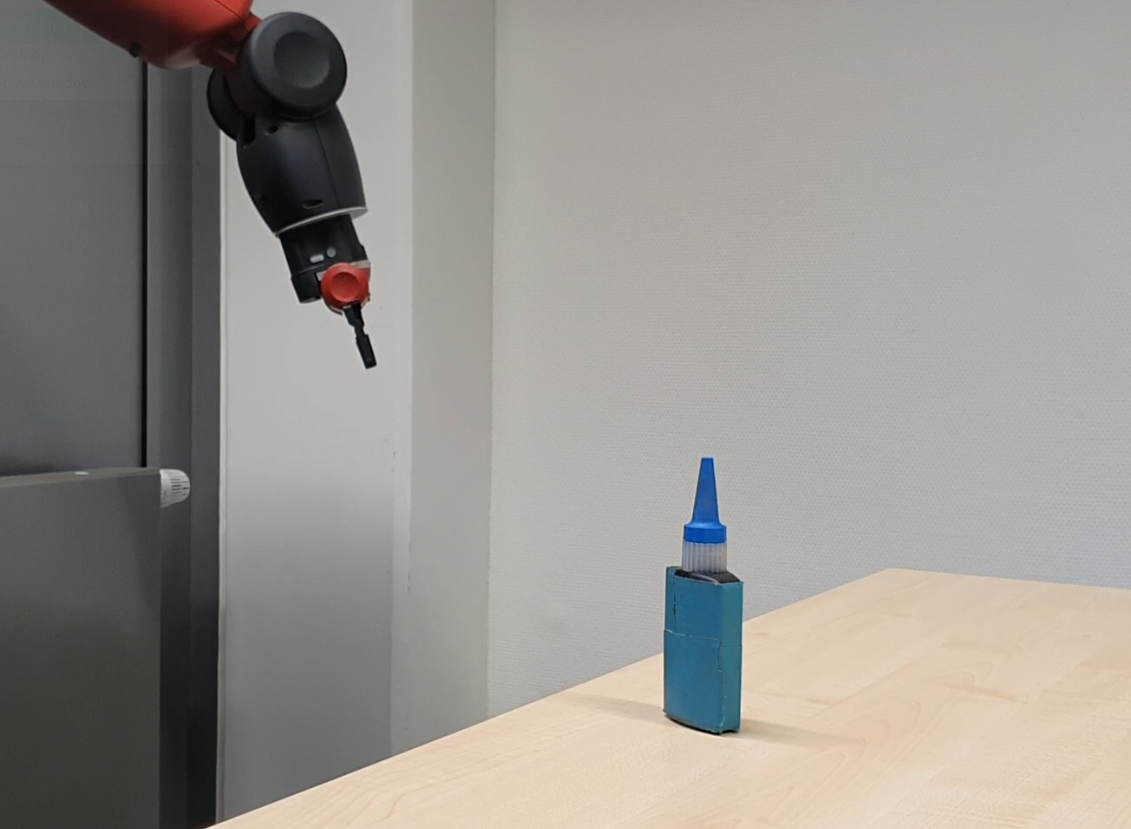}\hfill
	\hfill\hfill\includegraphics[width=3.0cm]{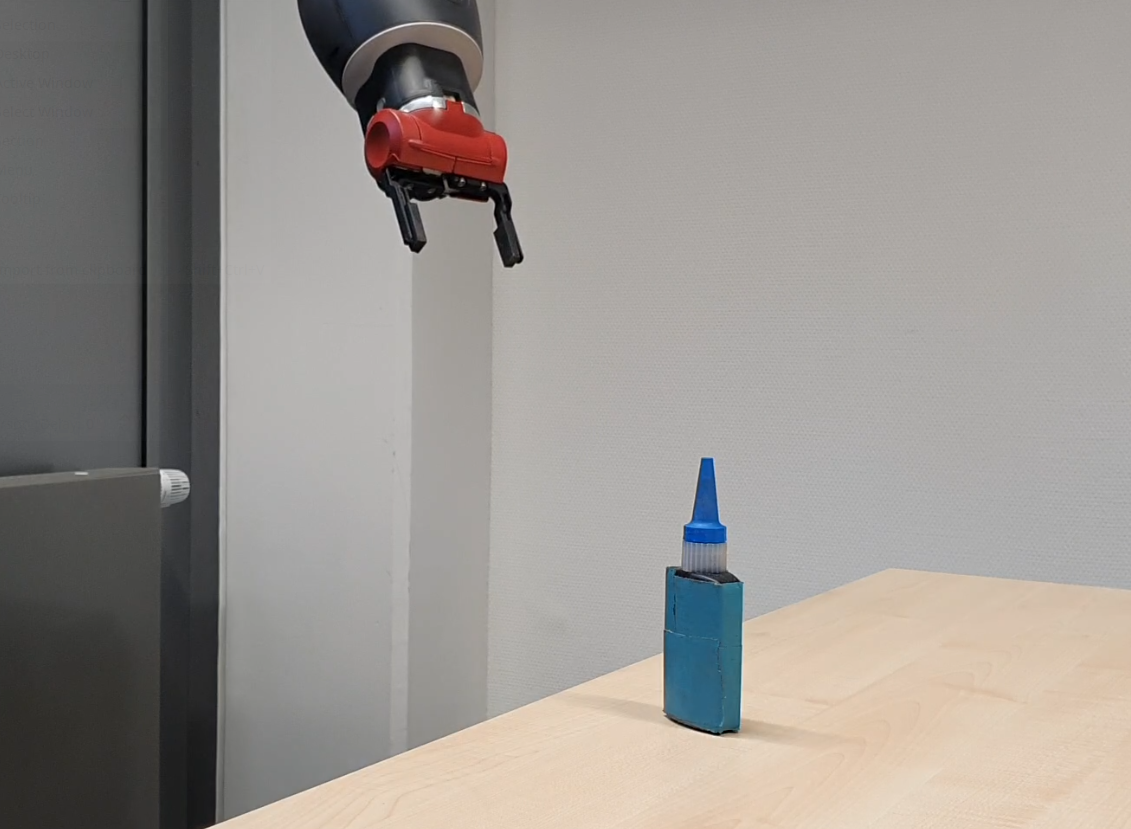}\hfill
	\hfill\hfill\includegraphics[width=3.0cm]{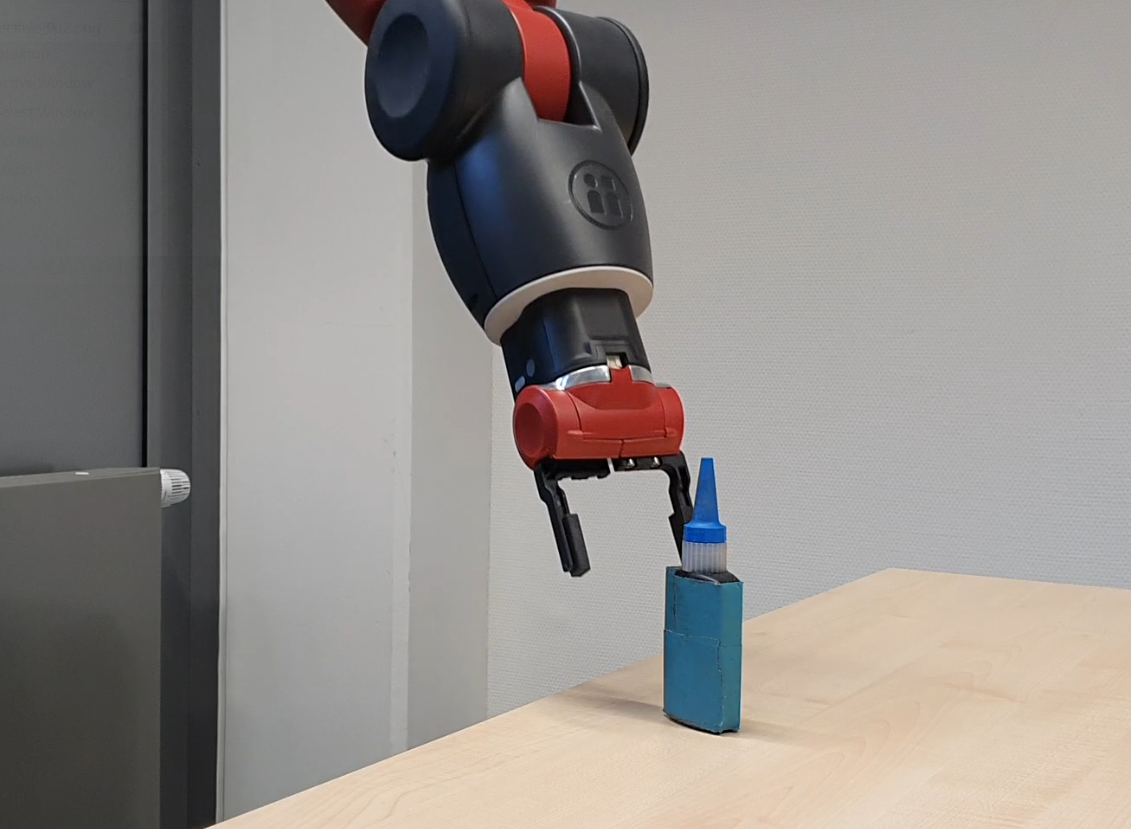}\hfill
	\hfill\hfill\includegraphics[width=3.0cm]{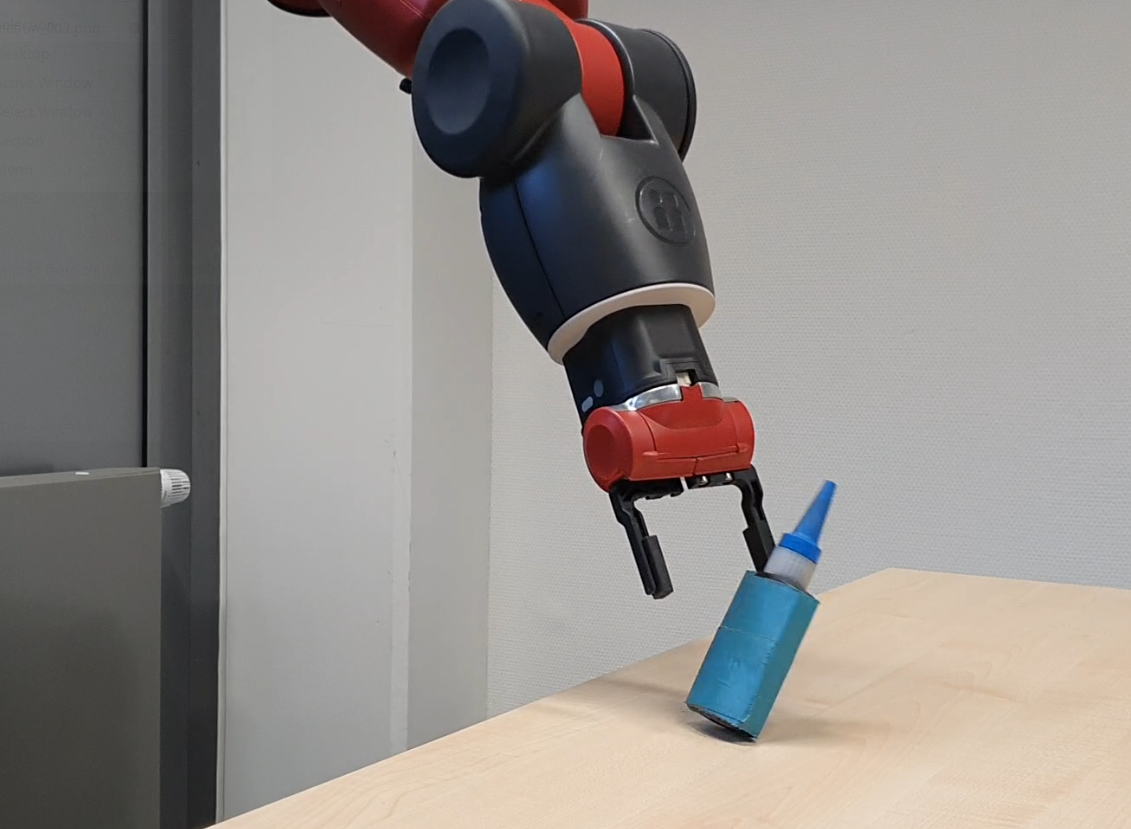}\hfill
	\hfill\hfill\includegraphics[width=3.0cm]{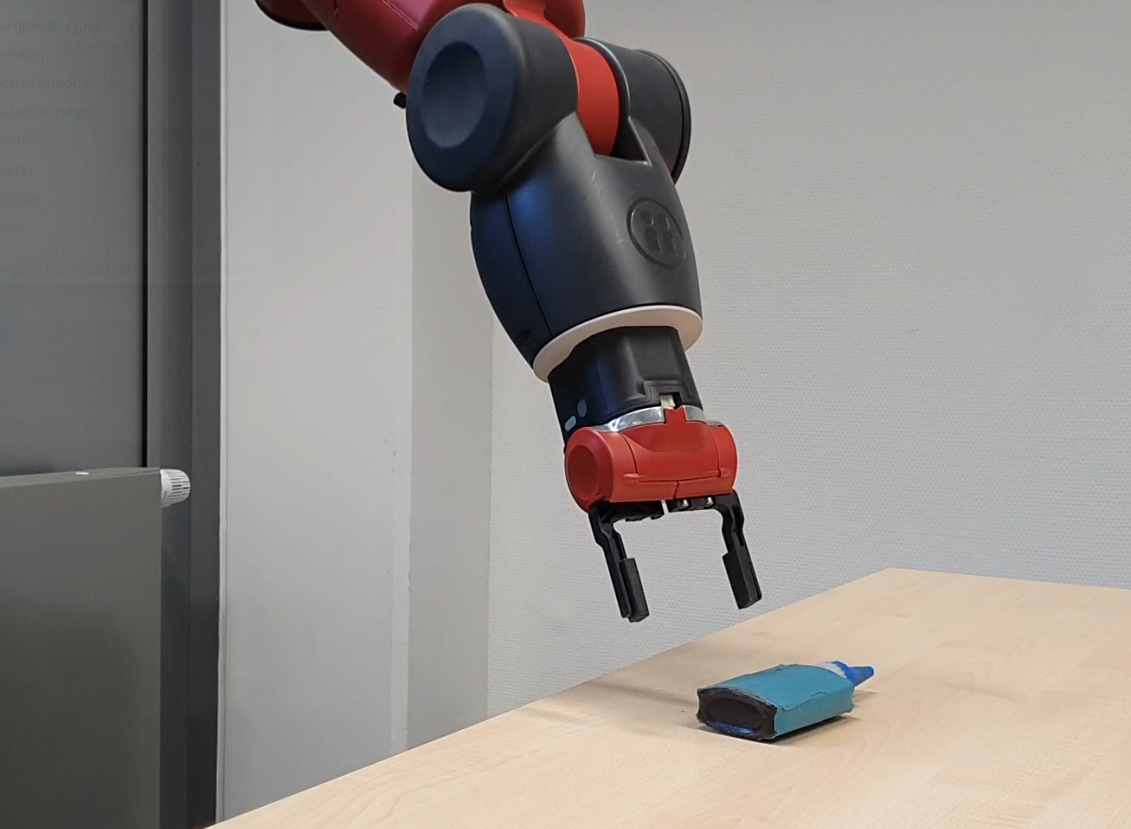}
	\\ \vspace*{1ex}
	a) \includegraphics[width=3.0cm]{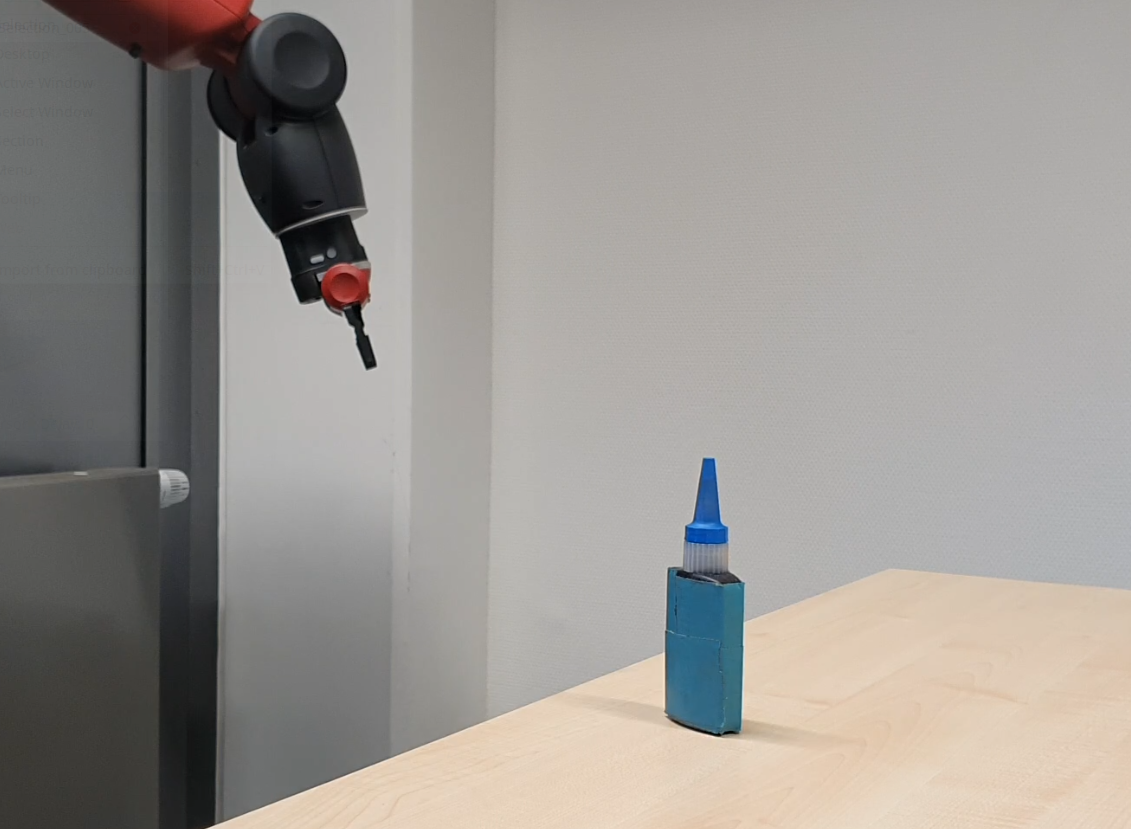}\hfill
	b) \includegraphics[width=3.0cm]{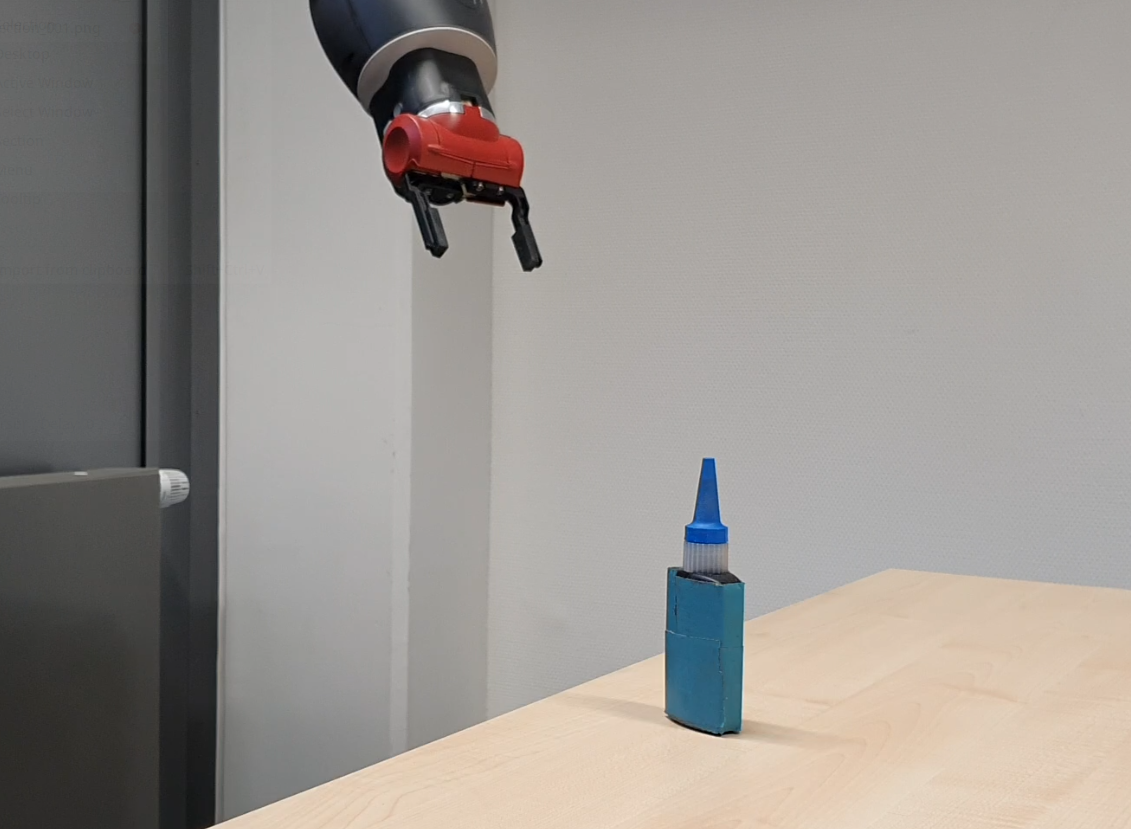}\hfill
	c) \includegraphics[width=3.0cm]{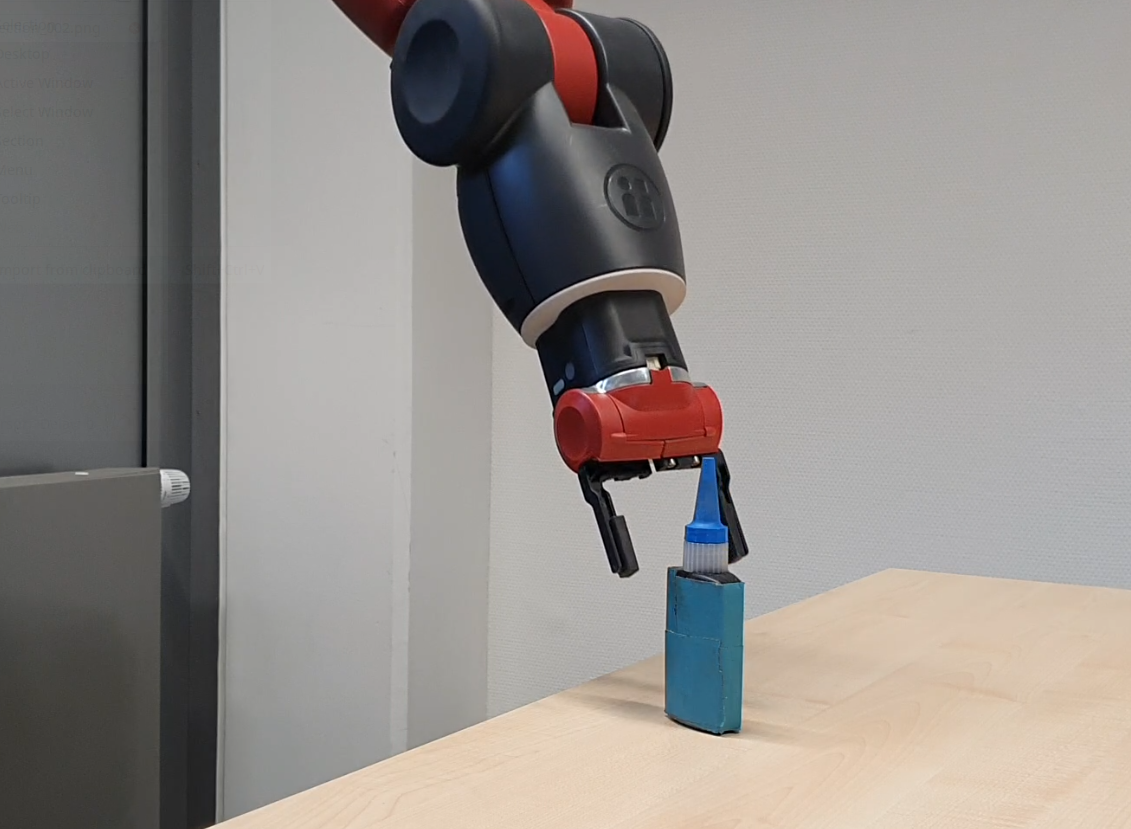}\hfill
	d) \includegraphics[width=3.0cm]{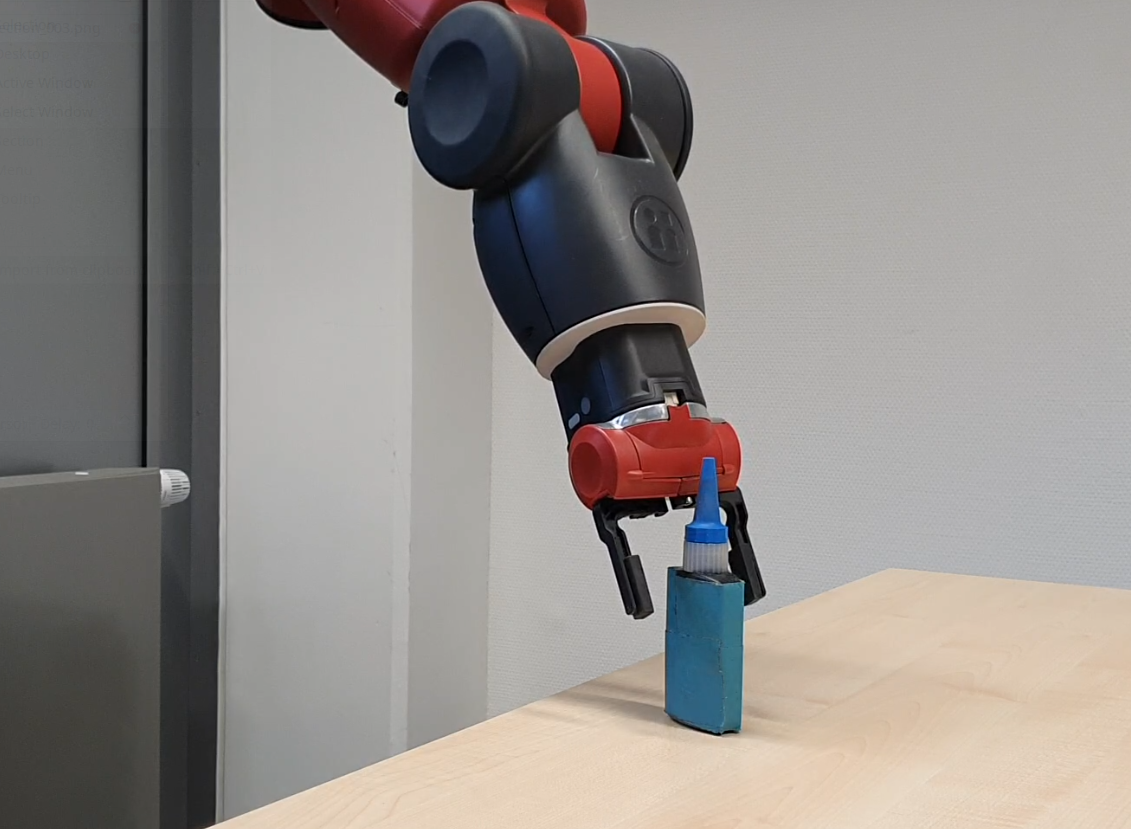}\hfill
	e) \includegraphics[width=3.0cm]{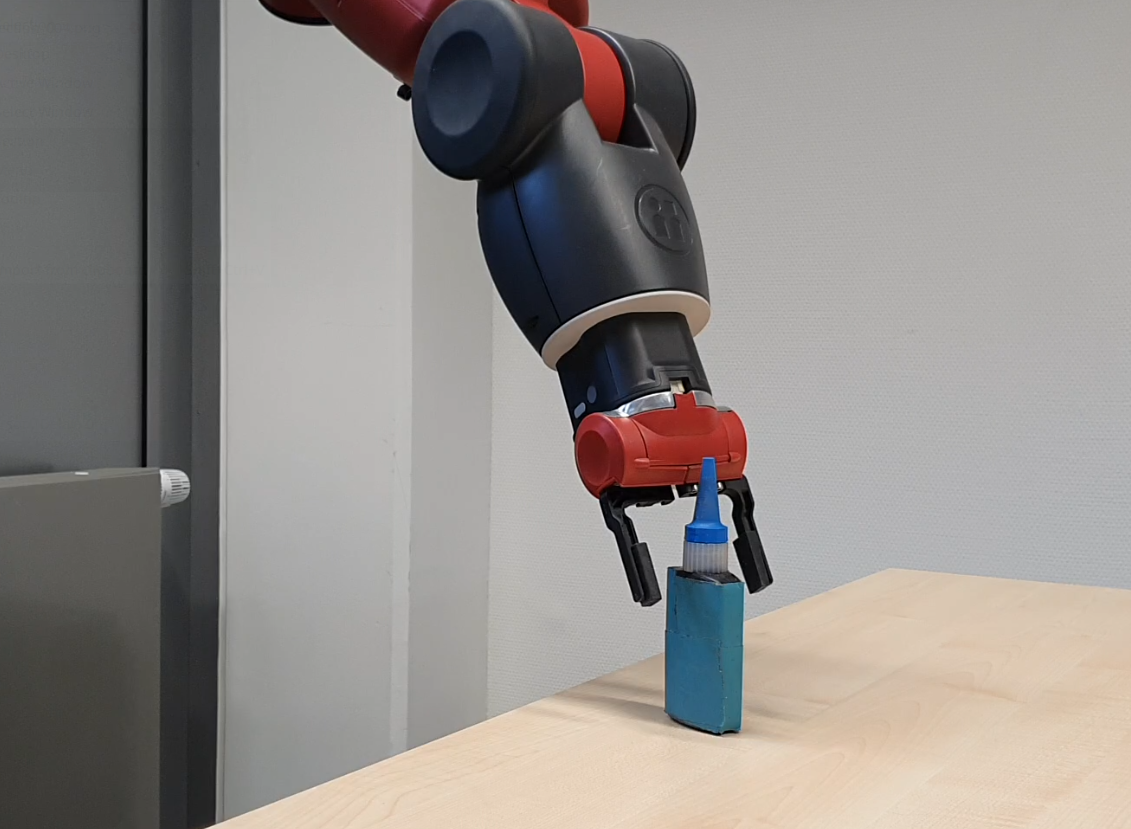}
	\caption{Pre-grasp trajectory execution. Top row: using only the baseline controller. Bottom row: using our model as an outer-loop controller. a) Start. b) Approach. c) Close proximity to the goal. In case of the baseline controller, inaccurate motion leads to a collision with the object. d-e) Pre-grasp pose reached. However, in case of the baseline controller, the collision led to tipping over the object. Our approach allowed to reach the goal position smoothly.}
	\label{fig:real_robot_pregrasp}
\end{figure*}

In \reffig{fig:3d_traj}, one can see that under the control of the baseline controller, the end-effector arrives to the goal pose in a curve, overshooting the desired path. Such behavior makes it very difficult to perform picking tasks with high speeds, because the end-effector often collides with objects at the pre-grasp pose. This often results in failed grasping attempts and can potentially damage the robot and its surroundings. A typical mitigation approach would be to define another pre-grasp pose and, upon arrival there, continue at a very low speed to the actual pre-grasp pose. This slows down the task execution, however. In this example, we demonstrate the effectiveness of our approach by reaching a pre-grasp pose at high speed. We execute the same trajectory first with the baseline controller and then with the proposed model as an outer loop controller. Two sequences of pictures in \reffig{fig:real_robot_pregrasp} show the execution of these trajectories. In Frame c) the baseline controller deviates from the path, leading to the collision with the object in Frame d). This results in the object being tipped over in Frame e). In contrast, our method tracks the trajectory more accurate and has a smoother velocity profile, which avoids the typical overshooting. This results in the successfully reached pre-grasp pose without colliding with the object. A video of the experiment is available on our website\footnote{\url{https://www.ais.uni-bonn.de/videos/IRC_2021_Pavlichenko}}.

\begin{table}
	\centering
	\caption{Comparison of average end-effector position error per point and average extra time to reach the endpoint.}
	\label{table:eef_error}
	\normalsize
	\begin{tabular}{lcc}
		\hline
		Method &
		\begin{tabular}[c]{@{}c@{}}EEF position error, cm\end{tabular} &
		\begin{tabular}[c]{@{}c@{}}Extra time, s\end{tabular} \\ \hline
		Baseline    & 2.981$\pm$0.355 ( --- ) & 0.75$\pm$0.091 ( --- ) \\
		FC          & 1.397$\pm$0.243 (53\%)  & 0.28$\pm$0.029 (62\%) \\
		RNN         & 1.151$\pm$0.142 (61\%)  & 0.12$\pm$0.013 (84\%) \\
		DWH         & 1.109$\pm$0.131 (62\%)  & 0.09$\pm$0.010 (88\%) \\
		FIN-DWH     & \textbf{1.015$\pm$0.122 (66\%)}  & \textbf{0.06$\pm$0.008 (92\%)} \\ \hline
	\end{tabular}
	\begin{tablenotes}
		\item \hspace{1ex} \footnotesize{95\% confidence interval is provided after "$\pm$". Improvement over the baseline is in brackets.}
	\end{tablenotes}
\end{table}

\begin{table}[b]
	\centering
	\caption{Comparison of average cumulative joint position and velocity error per point while carrying a payload.}
	\label{table:payload}
	\normalsize
	\begin{tabular}{clcc}
		\hline
		\begin{tabular}[c]{@{}c@{}}Payload,\\ kg\end{tabular} &
		\multicolumn{1}{c}{Method} &
		\begin{tabular}[c]{@{}c@{}}Joint position\\ error, rad\end{tabular} &
		\begin{tabular}[c]{@{}c@{}}Joint velocity\\ error, rad/s\end{tabular} \\ \hline
		\multirow{3}{*}{0.25} & Baseline                 & 0.074$\pm$0.014 & 0.341$\pm$0.064 \\
		& \multirow{2}{*}{FIN-DWH} & 0.039$\pm$0.008 & 0.217$\pm$0.035 \\
		&                          & (47\%)      & (36\%)      \\ \hline
		\multirow{3}{*}{0.5}  & Baseline                 & 0.080$\pm$0.016 & 0.347$\pm$0.067 \\
		& \multirow{2}{*}{FIN-DWH} & 0.042$\pm$0.009 & 0.226$\pm$0.038 \\
		&                          & (47\%)      & (34\%)      \\ \hline
		\multirow{3}{*}{1.1}  & Baseline                 & 0.108$\pm$0.019 & 0.378$\pm$0.083 \\
		& \multirow{2}{*}{FIN-DWH} & 0.071$\pm$0.013 & 0.256$\pm$0.046 \\
		&                          & (34\%)      & (32\%)      \\ \hline
	\end{tabular}
	\begin{tablenotes}
		\item \hspace{1ex} \footnotesize{95\% confidence interval is provided after "$\pm$". Improvement over the baseline is in brackets.}
	\end{tablenotes}
\end{table}

\subsection{Discussion}

\captionsetup[subfigure]{labelformat=empty}

The conducted experiments demonstrated that the proposed two-stage model architecture with one-step future prediction achieves a substantial improvement of the trajectory tracking accuracy, compared to the sole baseline controller, and outperforms the trivial models. The raw DWH model represents a reactive policy, which approximates the inverse dynamics of the manipulator. In contrast, the future prediction step of the FIN-DWH model allows to utilize the learned forward dynamics of the flexible-joint manipulator in order to find a corrective action, taking into account the predicted future inaccuracies. By making use of both forward and inverse dynamics, such an architecture allows extracting more useful knowledge from a significantly limited real-world dataset. The manipulators with flexible joints often do not have accurate dynamics models to perform training in simulation and/or have instance-specific dynamics properties due to wear and tear. Thus, the proposed method is useful for improving trajectory tracking of such robots. The achieved improvements do not come at a cost of increased stiffness and are based solely on learned dynamics of coupled joints.

One could argue that it is possible to develop a classical controller with higher accuracy of trajectory tracking than the Baxter baseline. However, this would require tedious tuning performed by an experienced professional on a time-span of several days. Moreover, such procedure is typically instance-specific and may need to be repeated after wear and tear accumulate during the use of the robot. At the same time, the proposed framework does not rely on any robot-specific parameters and requires only 45 minutes of data for the models to be trained. In addition, the learned models can be easily retrained to adjust for wear and tear, using the most recent recorded trajectories at any time, since the training procedure takes several hours on a regular computer. Finally, developing and tuning a classical controller which takes into account coupled joint dynamics is an extremely challenging task, which we resolve by following the data-driven approach.

The main limitation of the presented approach is its open-loop nature. Thus, any unexpected disturbances are unaccounted for, and remain for the inner-loop controller to be dealt with. As it was shown in the experiment with payloads, an increase of such unaccounted disturbance degrades the effectiveness of the method. Nevertheless, the model, learned by our method still achieves more accurate trajectory tracking, compared to the sole baseline controller, even with unaccounted disturbances. The underlying low-level classical controller provides guarantees on system stability. As the proposed method does not use feedback, it can be easily integrated into a software pipeline as a top-level addition over the existing low-level controller. Our methodology is agnostic of the type of the underlying classical controller and does not have robot-specific parameters, making it possible to apply it to most robotic manipulators without major changes. 
\section{Conclusion}	
\label{sec:Conclusion}

We presented a two-stage model based on linear time-invariant (LTI) dynamical operators for feed-forward outer loop control of a manipulator with flexible joints and unknown complex dynamics. The first part of the model estimates the future state of the system one step ahead with an unchanged control command. This estimation is used to augment the input for the second part of the model, which produces feed-forward joint position and velocity commands. The aim of this two-stage architecture is to push the model from reactive policy behavior towards more intelligent planning. The model was trained with backpropagation on a small 45\,min real-robot dataset. The approach was evaluated on the Baxter robot. Ablation study showed that one-step future prediction improved the performance. Our approach improved the trajectory tracking accuracy over the baseline controller: by 47\% and 35\% for the joint position and velocity tracking respectively, which resulted in 66\% improvement of the end-effector position tracking. This contributed to fast and smooth trajectory executions which required 92\% less extra time to reach the endpoint, allowing to perform tasks faster.

Future work includes exploring the possibility of employing a recurrent hierarchical model which is capable of looking several steps into the future, representing a model-predictive control approach. In addition, applying such a model in a closed-loop fashion in combination with online learning would allow obtaining a very flexible universal approach which would have the potential to further improve the trajectory tracking performance.

\printbibliography

\end{document}